# Solar Filaments Detection using Active Contours Without Edges

Sanmoy Bandyopadhyay* [(1)], and Vaibhav Pant [(1)]
(1) Aryabhatta Research Institute of Observational Sciences, Nainital, India-263002

## Abstract

In this article, an active contours without edges (ACWE)-based algorithm has been proposed for the detection of solar filaments in H$\alpha$ full-disk solar images. The overall algorithm consists of three main steps of image processing. These are image pre-processing, image segmentation, and image post-processing. Here in the work, contours are initialized on the solar image and allowed to deform based on the energy function. As soon as the contour reaches the boundary of the desired object, the energy function gets reduced, and the contour stops evolving. The proposed algorithm has been applied to few benchmark datasets and has been compared with the classical technique of object detection. The results analysis indicates that the proposed algorithm outperforms the results obtained using the existing classical algorithm of object detection.

## 1. Introduction

Solar filaments are one of the important features on the surface of the Sun. These filaments are large regions of very dense, cool gas, held in place by magnetic fields and generally sit above magnetic neutral lines, outlining the demarcation of negative and positive magnetic fields on the solar surface [1, 2]. Since they are cooler than their surroundings, they appear dark [1]. They usually appear long and thin above the chromosphere and are primarily observed in the H-alpha and Calcium K lines in the solar chromosphere. These solar features can be very dynamic objects, which change their appearance in only a few hours and last for a few weeks or months, looping hundreds of thousands of miles into space [3, 4].

If the magnetic field is destabilized, the filaments can erupt, e.g., as flares, coronal mass ejections (CMEs), and the plasma stored in them is ejected into space [4, 5, 6]. Every so often these filaments blow off from the surface of the Sun can cause space weather impacts on Earth in case their ways intersect [2]. These in turn raise the event of major interplanetary disturbances, and geomagnetic storms. Thus for better understanding of the changes in solar activities and operational space weather forecasts there arises the need to analyze the location, properties, and time evolution of solar filaments, which in turn requires the need for proper and automated identification of the filaments.

In literature there exists several automated methods of solar filaments detection. Gao *et al.* [7] used thresholding and region-based techniques to detect filaments in full-disk solar images. Candidate filament pixels were identified through thresholding, and a region growing method was applied to merge these pixels into filament areas [7]. Shih and Kowalski [8] proposed a method involving local and global thresholding to distinguish dark regions from the solar disk background. They applied directional morphological filtering to extract elongated shapes. Qu *et al.* [9] employed the stabilized inverse diffusion equation (SIDE) followed by adaptive thresholding for segmenting filament regions in solar images. Subsequently, a support vector machine (SVM) classifier was used on the segmented image to distinguish filaments from Sunspots. Atouma *et al.* [10] used adaptive local thresholding for filament segmentation. The segmented filaments were then evaluated by comparing the resulting image with a manually constructed synoptic map. Despite several advantages of thresholding-based methods, the selection of the optimal threshold value remains a challenging task [11]. Fuller and Aboudarham [12] introduced a filament detection method using seed selection and region growing technique. However, the success of the method depends on the proper selection of seeds point in the solar images. In recent decades, advancements in artificial neural network (ANN) and deep learning techniques have led to the adoption of methodologies such as fully convolutional neural network (FCN) u-net [13], Mask R-CNN [14], Conditional Convolutions for Instance Segmentation [15], and YOLOv5 [4] for filament identification. However, the success of these supervised neural network approaches hinges on rigorous training, necessitating extensive datasets for effective model performance. In the recent past K-means clustering technique had been used by Priyadarshi *et al.* [16] for the identification of the filaments. However the method had been applied on Kodaikanal Solar Observatory (KoSO) suncharts.

In order to overcome the aforementioned issue associated with the solar filaments detection and to detect the filaments region accurately, in this paper an active contour without edges (ACWE) [17] based algorithm has been introduced. The overall proposed algorithm of filament detection consists of three major steps of computer vision, these are image pre-processing, image segmentation and image post-processing. The major contribution in the paper is,

- Application of active contour without edges (ACWE) based technique for the detection of the filament region in full-disk solar image.

The proposed algorithm has been applied on a few benchmark Hα full-disk solar images captured in the year 2023. The resultant images have been compared with the results obtained from the classical methods of object detection.

The overall organization of the paper is as follows. In Section 2 the filaments detection using the proposed algorithm has been explained. Discussion on results has been carried out in Section 3. Finally, in Section 4 conclusions on the conducted work have been drawn.

## 2. Proposed Method of Filaments Detection

The proposed ACWE-based algorithm has been applied to Hα full-disk solar images for the detection of solar filaments. This detection process encounters two major issues. Firstly, the Hα solar images used in the experiment exhibit an inhomogeneous intensity distribution throughout, evident in the first row of Figure 1. Secondly, the ACWE method used in this detection process can be prone to local minima, potentially leading to false boundary detections. In order to address these issues in this work of filament detection three image pre-processing methodologies have been applied. In the first step image inpainting has been carried out, which involves filling in part of an image using information from the surrounding area [18]. The general equation of image inpainting can be written as,

$$I^{n+1}(x,y) = I^n(x,y) + \Delta t I_t^n(x,y), \forall (x,y) \in \Omega \quad (1)$$

where $n$ denotes the inpainting time, $(x,y)$ indicate image pixel coordinate, $\Delta t$ is the rate of improvement and $I_t^n(x,y)$ denotes the update of image $I^n(x,y)$ [19]. In the equation (1), the term $\Omega$ represents the region to be inpainted. The notion $I^{n+1}(x,y)$ denotes the improved version of $I^n(x,y)$, which is given by;

$$I_t^n(x,y) = \delta \vec{L}^n(x,y) \cdot \vec{N}^n(x,y), \quad (2)$$

where $\vec{L}^n(x,y)$ is the information to be propagated, $\vec{N}^n(x,y)$ denotes the propagation direction and $\delta \vec{L}^n(x,y)$ represents the change in the information. This image inpainting in this work has been done to remove the white patches from the solar image so that the contour doesn't get stuck around the white region in the image. In the second step, the inpainted image was enhanced using the logarithmic transformation, in order to map a narrow range of low pixel intensity input values to a wide range of resultant values [20]. Mathematically, the output image obtained after performing log transform is given by,

$$I_l(x,y) = r\log(1 + I^{n+1}(x,y)); \quad (3)$$

where $r$ is a scaling constant given by $255/\log(1 + I_{max}^{n+1}(x,y))$, and $I_{max}^{n+1}(x,y)$ is the maximum pixel value in the inpainted image. In the last stage of pre-processing, the enhanced image has been sharpened using the formulation given by [21],

$$I_S(x,y) = 5I_l(x,y) - [I_l(x+1,y) + I_l(x-1,y) + I_l(x,y+1) + I_l(x,y-1)]. \quad (4)$$

Now, with a view to segment out the filament regions from the pre-processed image ACWE technique has been applied. The ACWE algorithm focuses on finding the region of interest (RoI) boundary by utilizing an active contour approach. The active contour is dynamically evolved by minimizing the value of the energy function using the level set method. A contour is initialized within the boundary of the image and then evolves as the numerical partial differential equations are iteratively solved. Finally, it converges to the RoI boundary. The energy function in the ACWE model is given by:

$$F(c_1, c_2, C) = \mu \cdot Length(C) + \nu \cdot Area(inside(C)) + \lambda_1 \int_{inside(C)} |I_S(x,y) - c_1|^2 dxdy$$

$$+ \lambda_2 \int_{outside(C)} |I_S(x,y) - c_2|^2 dxdy, \quad (5)$$

where $C$ is the initialized curve, and the constants $c_1, c_2$ are the averages of $I_S$ inside and outside the curve $C$ respectively. The terms $\mu$, $\nu$, $\lambda_1$ and $\lambda_2$ are the weight parameters associated with the energy function. After segmenting the image using ACWE, post-processing tasks were performed. Connected components in the segmented image were analyzed using area-based morphological operations, and components failing to meet the threshold area value were removed. This step aimed to eliminate redundant regions identified during the segmentation process. The major advantage of using ACWE for detecting the filaments regions in solar image is that the method has the capability to detect and preserve the boundaries of the RoI, even in noisy images. [17].

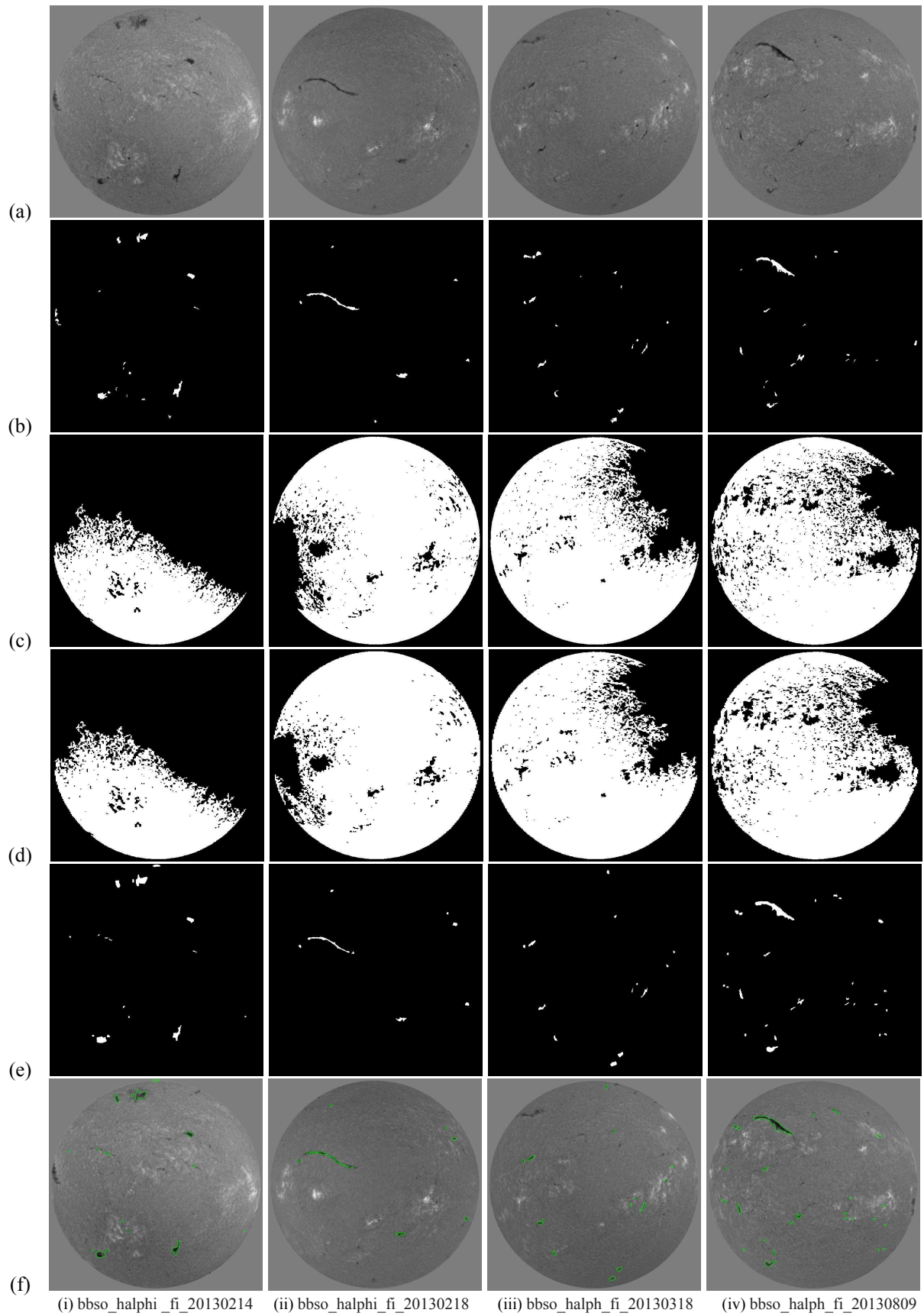

**Figure 1.** A visual summary compares classical algorithms—Otsu's thresholding (c) and K-means clustering (d)—with the proposed method (e) for filament detection in Hα full-disk solar images. The original image is depicted in (a), the ground-truth in (b), and the final result using ACWE in (f).

## 3. Results and Discussions

The experiment of filaments detection using the proposed algorithm were conducted on H$\alpha$ full-disk solar images captured by the Big Bear Solar Observatory (BBSO) during year 2013, dataset as used by Zhu et al. [13][1]. The dataset contain the ground-truth maps corresponds to the H$\alpha$ full-disk solar images, this in turn ease the visual and quantitative analysis of the performance of the proposed method. The parameter μ used in equation (5) is kept at 0.003, with a view to detect small as well as big filaments regions. The value assigned to the terms $\lambda_1$ and $\lambda_2$ are 1.000001 and 0.1 respectively. This has been done with a view to give more weight to the foreground regions in the image and extract out the dark filament regions as the background. The final output obtained using the proposed method is shown in the last row of Figure 1.

### 3.1 Visual Analysis

From the last two rows of Figure 1 it can be visualized that the proposed method is capable of detecting most of the filaments regions properly compared to the classical Otsu thresholding and K-mean clustering based methods of detection. In this case the classical method of detection fails because it tries to find out the filament by separating foreground and background class based on the optimal intensity value which mainly gets stuck near the average value of image pixel intensities. At the same time while comparing the results with the ground-truth images, it can be seen from Figure 1., that the proposed method has either identified a few small redundant regions as filaments or is unable to detect the filaments present near the solar boundary. This is due to the limitation of ACWE of getting stuck at the local minima.

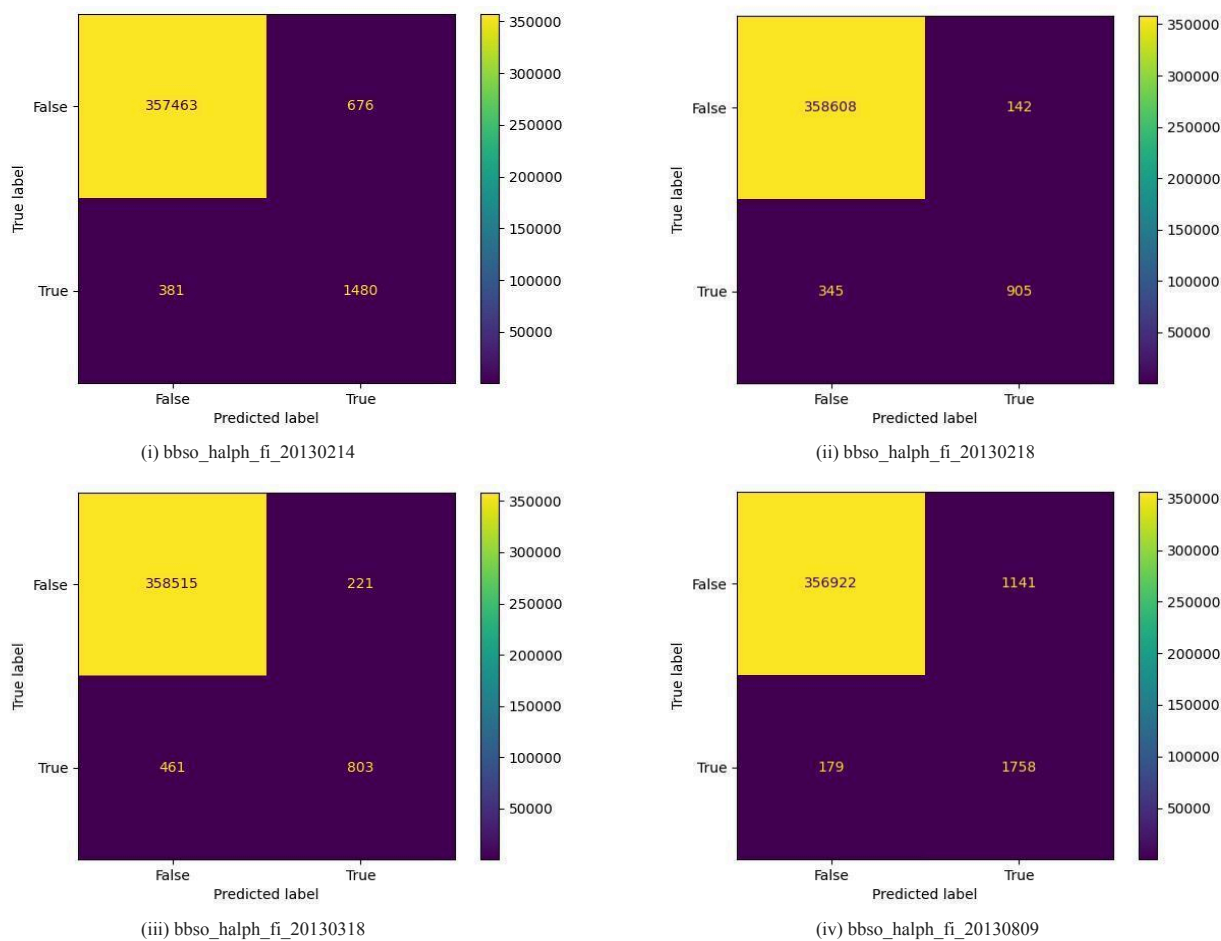

(i) bbso_halph_fi_20130214

(ii) bbso_halph_fi_20130218

(iii) bbso_halph_fi_20130318

(iv) bbso_halph_fi_20130809

**Figure 2.** Confusion matrix highlighting the performance of the proposed method in detecting the filaments in H$\alpha$ full - disk solar images.

### 3.2 Quantitative Analysis

---

[1] https://sun.bao.ac.cn/hsos_data/download/filaments-unet-dataset/

The performance of the proposed method has been quantitatively analyzed using the confusion matrix, depicted in Figure 3. From the matrix values, accuracy rate (AR) and true positive rate (TPR) were calculated, revealing that the proposed method achieves a maximum TPR of 0.9075 for the solar image captured on 2013-08-09. Meanwhile, classical methods show a TPR of 1.0. Interestingly, when comparing AR, the proposed method achieves 99.63%, significantly higher than the classical methods' AR of approximately 40%. This discrepancy arises because classical methods tend to misclassify redundant regions as filaments during segmentation.

## 4. Conclusion

In this study, an active contour algorithm without edges was implemented for detecting filament regions in H$\alpha$ full-disk solar images. Preprocessing techniques such as image inpainting, log transformation, and sharpening were applied to prepare the images for segmentation using ACWE. The proposed algorithm has notable advantages: it accurately identifies RoI boundaries even in noisy images and operates effectively without requiring a large image database like deep learning approaches do. However, the execution time of the algorithm (near about 29.8128 sec) is higher than the compared algorithms of detection. Future efforts will concentrate on enhancing its computational efficiency, potentially by refining the preprocessing stages. Additionally, aim to analyze solar filaments in extreme ultraviolet images from observatories like Aditya-L1 and Solar Dynamics Observatory in future.

## 5. Acknowledgements

The financial support received under Aditya-L1 Science Support Cell grant is thankfully acknowledged.